%% file: main.tex
\documentclass[letterpaper]{article} % DO NOT CHANGE THIS
\usepackage[draft]{aaai25}
\usepackage{times}  % DO NOT CHANGE THIS
\usepackage{helvet}  % DO NOT CHANGE THIS
\usepackage{courier}  % DO NOT CHANGE THIS
\usepackage[hyphens]{url}  % DO NOT CHANGE THIS
\usepackage{graphicx} % DO NOT CHANGE THIS

\usepackage{natbib}  % DO NOT CHANGE THIS AND DO NOT ADD ANY OPTIONS TO IT
\usepackage{caption} % DO NOT CHANGE THIS AND DO NOT ADD ANY OPTIONS TO IT
\usepackage{algorithm}
\usepackage{algorithmic}

\usepackage{booktabs} % For professional-looking tables
\usepackage{multirow}
\usepackage{xcolor}
\usepackage{colortbl}
\usepackage{tikz}

\definecolor{keywords}{RGB}{255,0,90}
\definecolor{comments}{RGB}{0,0,113}
\definecolor{red}{RGB}{160,0,0}
\definecolor{green}{RGB}{0,150,0}

\definecolor{bananayellow}{rgb}{1.0, 0.88, 0.21}

\usepackage{amsmath}

\usepackage{newfloat}
\usepackage{listings}
\DeclareCaptionStyle{ruled}{labelfont=normalfont,labelsep=colon,strut=off} % DO NOT CHANGE THIS
\floatstyle{ruled}
\newfloat{listing}{tb}{lst}{}
\floatname{listing}{Listing}
\title{Knowledge Tagging with Large Language Model based Multi-Agent System}
\author {
    Hang Li\textsuperscript{\rm 1,2},
    Tianlong Xu\textsuperscript{\rm 1},
    Ethan Chang\textsuperscript{\rm 3},
    Qingsong Wen\textsuperscript{\rm 1}\thanks{Corresponding author}
}
\affiliations {
    \textsuperscript{\rm 1}Squirrel AI Learning, Bellevue, WA, USA\\
    \textsuperscript{\rm 2}Michigan State University, East Lansing, MI, USA \\
    \textsuperscript{\rm 3}Columbia University, New York, NY, USA \\
    lihang4@msu.edu, \{tianlongxu,qingsongwen\}@squirrelai.com, ethan.chang@columbia.edu
}
\usepackage{bibentry}
\begin{document}

\maketitle

\begin{abstract}
Knowledge tagging for questions is vital in modern intelligent educational applications, including learning progress diagnosis, practice question recommendations, and course content organization. Traditionally, these annotations have been performed by pedagogical experts, as the task demands not only a deep semantic understanding of question stems and knowledge definitions but also a strong ability to link problem-solving logic with relevant knowledge concepts. With the advent of advanced natural language processing (NLP) algorithms, such as pre-trained language models and large language models (LLMs), pioneering studies have explored automating the knowledge tagging process using various machine learning models. In this paper, we investigate the use of a multi-agent system to address the limitations of previous algorithms, particularly in handling complex cases involving intricate knowledge definitions and strict numerical constraints. By demonstrating its superior performance on the publicly available math question knowledge tagging dataset, MathKnowCT, we highlight the significant potential of an LLM-based multi-agent system in overcoming the challenges that previous methods have encountered. Finally, through an in-depth discussion of the implications of automating knowledge tagging, we underscore the promising results of deploying LLM-based algorithms in educational contexts.

\end{abstract}

\section{Introduction}
\label{sec:intro}

\input{introduction}

\section{Related Work}

\input{related}

\section{Problem Statement}

\input{problem}

\section{System Design}

\input{method}

\section{Experiment}

\input{experiment}

\section{Industrial Impact}

\input{impact}

\section{Conclusion}

In this paper, we introduce a novel LLM-based multi-agent framework for the knowledge-tagging task, which leverages the "divide and conquer" problem-solving strategy to address the complex cases involving intricate knowledge definitions and strict numerical constraints that have challenged previous algorithms. Through the precise collaboration of diverse LLM agents, our system harnesses the strengths of individual agents while integrating external tools, such as Python programs, to compensate for LLMs' limitations in numerical operations. To validate the effectiveness of the proposed framework, we conducted experiments using the expertly annotated knowledge concept question dataset, MathKnowCT. The results demonstrate the framework's efficacy in enhancing the knowledge-tagging process. Finally, through a detailed discussion of the implications of automating knowledge tagging, we highlight the promising future of deploying LLM-based algorithms in educational contexts.

\bibliography{aaai25.bib}

\end{document}

%% file: introduction.tex
\begin{figure*}[!btph]
    \centering
    \includegraphics[width=0.99\textwidth]{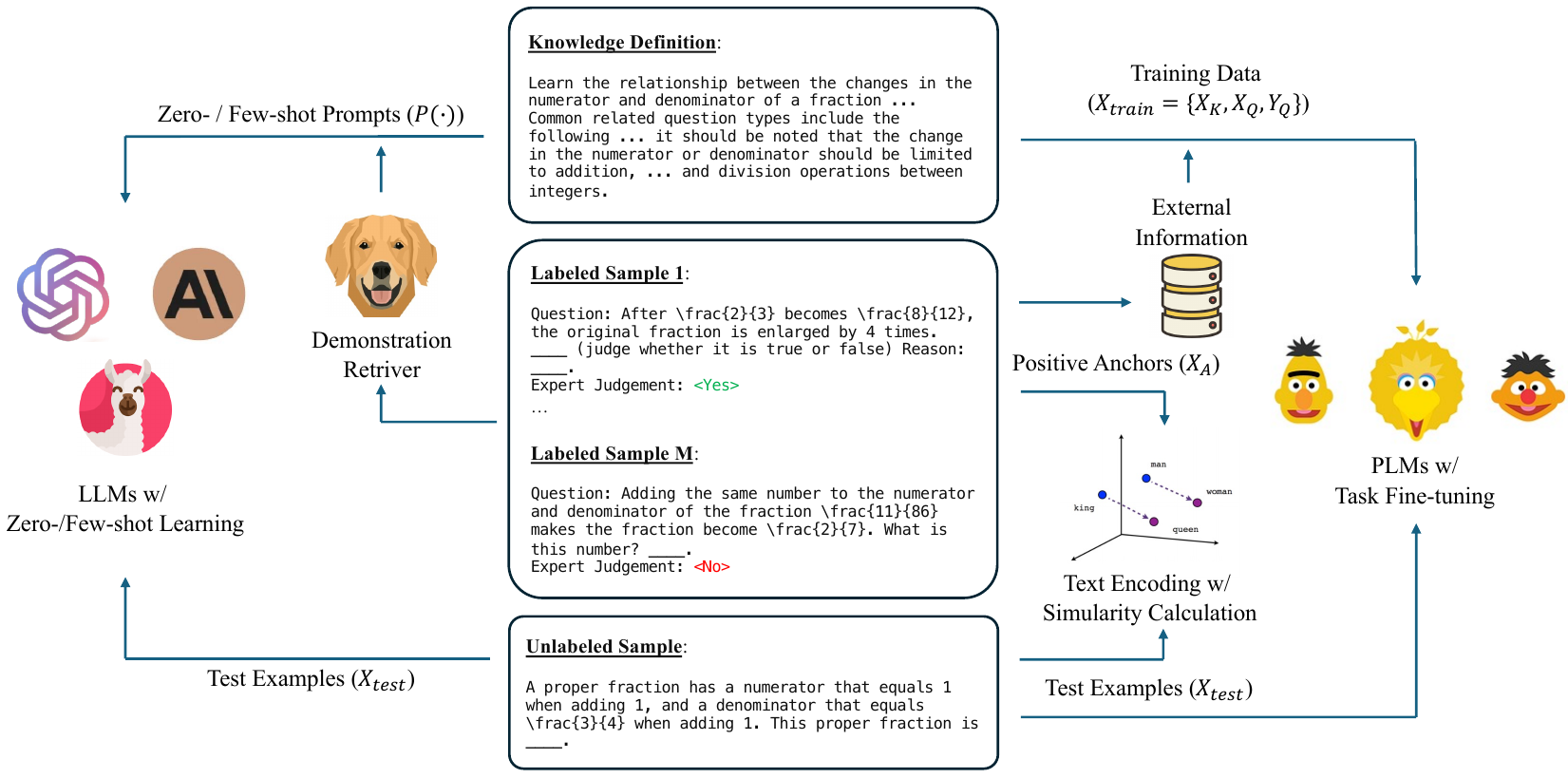}
    \caption{An summary of the existing algorithm for automatic tagging task.} 
    \label{fig:summary}
\end{figure*}

Knowledge tagging is focused on creating an accurate index for educational content. It has become a key element in today's intelligent education systems, essential for delivering high-quality resources to educators and students~\citep{chen2014tag}. For example, with well-tagged educational materials, teachers can easily organize course content by searching through a concept keyword index~\citep{sun2018automatic}. Traditionally, educational experts have manually annotated concept tags for questions. However, the rapid expansion of online content has made these manual methods insufficient to keep up with the growing volume of online question data and the need to update concept tags quickly~\citep{li2024bringing}. To solve the above issues, recent studies have tried to automate the tagging process with different natural language processing (NLP) algorithms~\cite{wang2024large}. For instance, \citet{sun2018automatic} employ deep learning algorithms and coverts the tagging task as a binary classification problem. Other works~\citep{huang2023pqsct} fuse external information, e.g., solution text and conceptual ontology, with original question contents during the judging process. The most recent work~\citep{li2024knowledge} leverages  large language models (LLMs) and simulates human expert tagging process with the helps on chain-of-thought (COT)~\citep{wei2022chain} and in-context learning (ICL)~\citep{dong2022survey} tricks during the automatic knowledge tagging process. In the Fig.~\ref{fig:summary}, we summarize the existing algorithm for automatic knowledge tagging task. Although, all these studies have demonstrated appealing results in their experiments, there are still limitations in each algorithm causing gaps between human performance and automatic tagging results. 

In this work, we propose a novel LLM-based multi-agent system (MAS) for knowledge tagging task, which exploits the planning and tool-using capabilities on LLMs. Specifically, by reformulating the judging process into a collaboration between multiple LLM-agents on independent sub-problems, we simplify the whole task and enhance the reliability of the judgment generation process. To validate the effectiveness of our proposed algorithm, we experiment with a well-established knowledge concept question dataset MathKnowCT~\citep{li2024knowledge}. Our experimental results demonstrate that our method can bring steady improvements to prior single LLM-based methods. 

%% file: related.tex
\subsection{Knowledge Tagging}

The recent rapid advancements in the field of machine learning (ML) have encouraged the emergence of studies focused on applying advanced ML models to address challenging problems in education~\citep{xu2024foundation,wang2020neural}. One critical area of exploration is the automatic knowledge tagging task, which is essential for modern Intelligent Tutoring Systems (ITS). \citet{sun2018automatic} were among the first to utilize straightforward models like long short-term memory (LSTM) networks and attention mechanisms to learn short-range dependency embeddings. In their approach, questions are processed through neural network layers and linked to cross-entropy functions to determine if a tagging concept is relevant to a specific problem. Building on this, \citet{liu2019ekt} designed an exercise-enhanced recurrent neural network with Markov properties and an attention mechanism to extract detailed knowledge concept information from the content of exercises. Similarly, enriched data sources such as text, multi-modal data~\citep{yin2019quesnet}, and combined LaTeX formulas~\citep{huang2021context} have been used to improve semantic representations learned with LSTM, allowing for the capture of more implicit contexts. To leverage the robust transformers framework, \citet{zemlyanskiy2021docent} pretrained a BERT model to jointly predict words and entities as movie tags based on movie reviews. \citet{huang2023pqsct} introduced an enhanced pretrained bidirectional encoder representation from transformers (BERT) for concept tagging, incorporating both questions and solutions. With the rise of large language models (LLMs), recent pioneering studies~\citep{li2024knowledge} have explored using LLMs as evaluators, simulating the human expert tagging process with the aid of chain-of-thought (COT) and in-context learning (ICL) techniques. LLM-based algorithms offer significant advantages in handling cases where annotation samples are scarce or unavailable, leveraging their extensive prior knowledge.

\subsection{Multi-Agent System}

An LLM-based multi-agent system (MAS) incorporating large language models (LLMs) consists of multiple autonomous agents, each potentially utilizing LLMs, working together to achieve particular goals~\citep{guo2024large}. These systems take advantage of LLMs to boost the agents' capabilities, intelligence, and adaptability. MAS generally includes three key components: agents, communication protocols, and coordination mechanisms. The agents, driven by LLMs, are tasked with executing actions and are initiated by specific role-prompts tailored to individual tasks, such as programming, answering queries, or strategic planning. Communication protocols establish how agents share information, often using natural language conversations, structured message exchanges, or other methods of interaction. Coordination mechanisms are vital in MAS, as they handle the complexity and independence of each agent. When applied to education, LLM-based MAS has presented its great potentials in various piratical usages~\citep{wang2024large}. For example, \citet{zhang2024simulating} simulate instructional processes by enabling role-play interactions between teacher and student agents with MAS. By analyzing agent behavior against real classroom activities observed in human students, studies have demonstrated that these interactions closely resemble real-life classrooms and foster effective learning. Beyond simulation, MAS has also been employed to improve LLM performance in tasks such as grading assignments~\citep{lagakis2024evaai} and identifying pedagogical concepts~\citep{yang2024content}. The introduction of multiple judging agents in group discussions has led to evaluations that align more closely with expert annotations.

%% file: problem.tex
Following the successful experience of applying LLMs for knowledge tagging task with ICL method in prior work~\citep{li2024knowledge}, we define the knowledge tagging problem as follows: given a pair of knowledge definition text $k$ and a question's stem text $q$, the objective of a concept tagging model $\mathcal{F}$ is to produce a binary judgment $y\in\{0,1\}$, where $\mathcal{F}(k,q) = 1$ means $k$ and $q$ are matching, 0 otherwise.

%% file: method.tex
\begin{figure*}[!btph]
    \centering
    \includegraphics[width=0.99\textwidth]{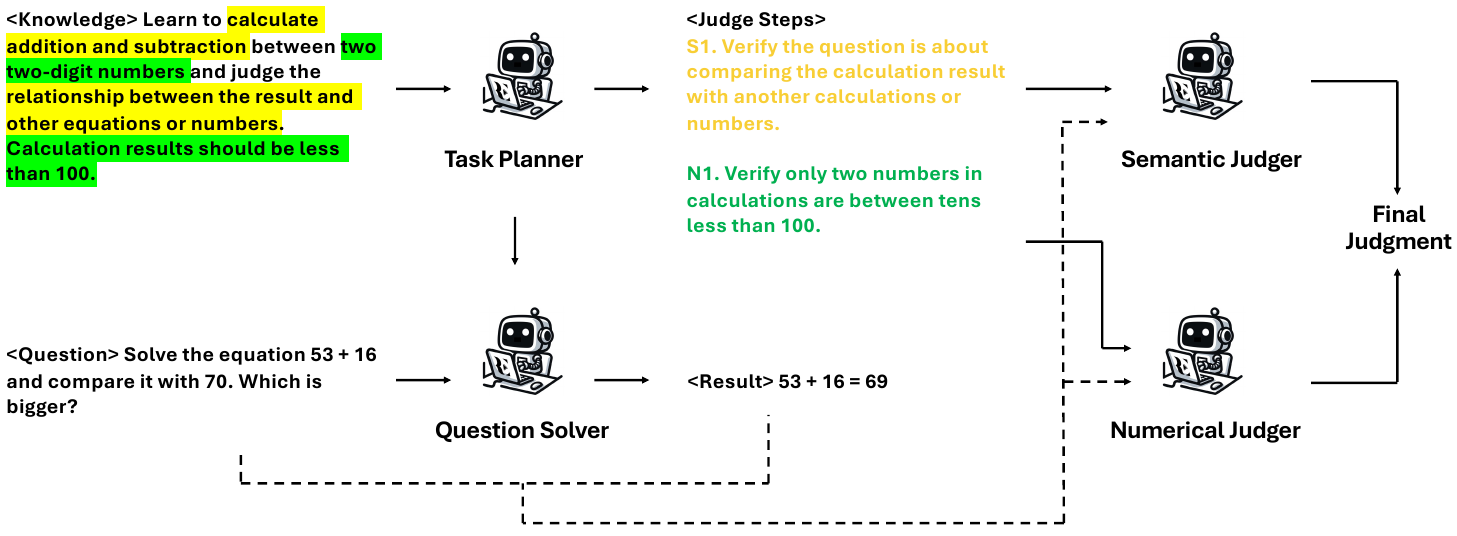}
    \caption{An overview of the proposed LLM-based multi-agent system for knowledge tagging. The {\color{bananayellow} \textbf{semantic}} and {\color{green} \textbf{numerical}} constraints in knowledge definition and decomposed sub-tasks are marked with corresponding colors.} 
    \label{fig:framework}
\end{figure*}

In this section, we introduce our LLM-based MAS. We first give an overview of the framework. Then, we detail the instructional prompts and other implementations details of different LLM-agent in independent sections.

\subsection{An Overview}

Our LLM-based MAS consists of four types of LLM-based agents, including task planner, question solver, semantic judger, numerical judger. At the beginning of the judging pipeline, the planning agent is activated proposing customized collaborating plan for the given knowledge definition. Then, the remaining agents are executed based the proposed plan. At last, the summarizing module outputs the final judgement by connecting those intermediate results with the AND operator. Fig.~\ref{fig:framework} presents an overview of the MAS and its workflow for the knowledge annotation task.

\subsection{Task Planner}

Knowledge concept definition is commonly composed by two major components: descriptive text and additional constrains. The duty of the task planner is to decompose the original definition into a series of independent verification sub-tasks and assign these tasks to the following agents. In general, by executing the step-by-step checking procedure, we avoid asking LLMs to proceed with multiple constraints at once, as it simplifies the task and helps the annotating system to generate accurate final judgments. In Fig.~\ref{fig:framework}, we present an example plan for the given knowledge concept. Based on the knowledge description, the planner proposes four sub-tasks, including 1 semantic judge and 3 numerical judges. The prompt for the planning agent is shown below, where the [Example] is the placeholder for implementing the few-shot learning tricks~\citep{brown2020language}.

\begin{quote}

\emph{\textbf{Instruction}}: Your job is to take the following knowledge definition and separate it into one or more simpler Knowledge sub-constraints. Each of these smaller constraints must return Yes or No value when evaluated. examples: [Example 1] [Example 2]

\emph{\textbf{Knowledge}:} [Input by User]

\emph{\textbf{Plan}:} (Generated by LLMs)

\end{quote}

\subsection{Question Solver}

In addition to the planner, we have integrated a question solver into the system to generate solutions for questions where constraints in knowledge definitions may impact the solution values. Since question-solving tasks are widely utilized by all LLMs during the instructional tuning phase, we do not employ any additional engineering techniques. Instead, we compose the prompt for the agent as follows:

\begin{quote}

\emph{\textbf{Instruction}}: You are a student. Given a question, provide the answer at the end.

\emph{\textbf{Question}:} [Input by User]

\emph{\textbf{Answer}:} (Generated by LLMs)

\end{quote}

\subsection{Semantic Judger}

The semantic judger is designed to execute verification tasks based on the semantic constraints outlined in the knowledge definition. Leveraging the general prior knowledge of LLMs, the LLM-based agent is adept at understanding semantic patterns between the input knowledge and question pairs. In our implementation, we employ standard sequential generation and incorporate few-shot learning to further enhance performance. The detailed instruction prompt used for the semantic judger is as follows:

\begin{quote}

\emph{\textbf{Instruction}}: You are a knowledge concept annotator. Your job is to judge whether the question is concerning the knowledge. The judgment tokens $<$Yes$>$ or $<$No$>$ should be provided at the end of the response. You are also given two examples. [Example 1] [Example 2]

\emph{\textbf{Knowledge}:} [Copied from Task Planner]

\emph{\textbf{Question}:} [Copied from Question Solver]

\emph{\textbf{Judgment}:} (Generated by LLMs)

\end{quote}

\subsection{Numerical Judger}

Although LLMs excel at handling semantic-related instructions, recent studies~\citep{collins2024evaluating} have shown that they struggle with numerically related requests when relying solely on a sequential generation strategy. To address this issue, we draw inspiration from the recently emerged Tool-use LLMs~\citep{zhuang2024toolqa} and leverage the LLMs' emergent coding capabilities to verify constraints through code execution. Specifically, we process the numerical judging procedure in two steps. First, the LLM extracts relevant numbers from the question stem to use as arguments in a Python program. Then, the LLM is instructed to convert constraints into executable code. Below, we present an example of a numerical judging prompting process for a given knowledge and question pair.

\begin{quote}

\emph{\textbf{Instruction1}}: You are a knowledge concept classifier. You are given a Question, Answer, a Main constraint, and Subconstraints. Identify the numerical arguments for each subconstraint. [example]

\emph{\textbf{Knowledge}:} [Copied from Task Planner]

\emph{\textbf{Question}:} [Copied from Question Solver]

\emph{\textbf{Sub-constraints}:} [Copied from Task Planner]

\emph{\textbf{Argument}:} (Generated by LLMs)

\emph{\textbf{Instruction2}}: You are a knowledge concept annotator. You are given a Question, Answer, Sub-contraints, and Arguments. Your job is to write a Python script using the sub-constraints and their respective arguments and evaluate them. The script prints True if all the sub-constraints return True, False if else.

\emph{\textbf{Knowledge}:} [Copied from last step]

\emph{\textbf{Question}:} [Copied from last step]

\emph{\textbf{Argument}:} [Copied from last step]

\emph{\textbf{Sub-constraints}:} [Copied from Question Solver]

\emph{\textbf{Program Code}:} (Generated by LLMs)

\end{quote}

\noindent Once the executable program code is generated, the agent automatically runs the program with all relevant arguments. The final judgment is then determined by evaluating the program's boolean output.

\begin{figure}[!btph]
\centering
\begin{tikzpicture}
  \draw node[draw=black,fill=black!20,rounded corners,inner sep=2ex,text width=0.44\textwidth] {
    \textbf{Knowledge:} 
    
    Learn to perform addition and subtraction operations between two tens less than 100, and judge the relationship between the results of the calculation and other calculations or numbers.
    
    \textbf{Question:} 
    
    Calculate the following equation: 20 + 50 and compare the result with 90, which is larger?
    
    \textbf{Task Planner:} 
    
    \begin{itemize}
        \item N1: Verify only two numbers in calculations are between tens less than 100. 
        \item S1: Verify the question is about comparing the calculation result with another calculations or numbers.  
    \end{itemize}

    \textbf{Question Solver:} 
    
    \textit{Skipped} (Plans is not applicable to solutions.)

    \textbf{Numerical Judger (Step 1):} 
    
    Extract input argument from the question: 20 + 50.
    
    \textbf{Numerical Judger (Step 2):}    
    \lstset{language=Python, 
            basicstyle=\ttfamily\small, 
            keywordstyle=\color{keywords},
            commentstyle=\color{comments},
            stringstyle=\color{red},
            showstringspaces=false,
            identifierstyle=\color{green},
            keywords=[2]{pow},
            keywordstyle=[2]{\color{orange}},
    }
    
    \begin{lstlisting}[xleftmargin=0pt, numbers=none]
import re

def verify_add_subtract_calculation(input_string):
    # Regular expression to match an addition or subtraction calculation between two numbers
    pattern = r'^\s*(\d{1,2})\s*([\+\-])\s*(\d{1,2})\s*$'
    
    match = re.match(pattern, input_string)
    
    if match:
        num1, operator, num2 = match.groups()
        num1, num2 = int(num1), int(num2)
        
        # Check if both numbers are tens and less than 100
        if num1 % 10 == 0 and num2 % 10 == 0 and num1 < 100 and num2 < 100:
            return True
        else:
            return False
    else:
        return False

print(verify_add_subtract_calculation('20 + 50'))
    \end{lstlisting}
    
     The output returned by python program is [python execution]. Thus, $<$Yes$>$.
    
    \textbf{Semantic Judger:} 
    
    The question asks to first calculate an equation and then compares it with a given number. This is aligned with given knowledge definition comparing the calculation result with numbers. Thus, $<$Yes$>$.

    \textbf{Final Judgment:} $<$Yes$>$
    };
\end{tikzpicture}
\caption{An example of step-wise outputs from different LLM-based agents.}
\label{fig:example}
\end{figure}

%% file: experiment.tex
In this section, we conduct experiments to validate the effectiveness of our purposed system. Through the experiments, we aim to answer the following research questions:

\begin{itemize}
  \item RQ1: Does the proposed method outperform the other baseline algorithms?
  \item RQ2: In which scenario, the proposed method shows it advantages?
\end{itemize}

\begin{table*}
\centering
\caption{Detailed sample statistics for different knowledge concepts in MathKnowCT.}
\label{tab:data_detail}
\resizebox{0.85\textwidth}{!}{
\begin{tabular}{@{}cccc|cccc@{}}
\toprule
\textbf{Knowledge ID} & \textbf{Total Size} & \textbf{Positive Size} & \textbf{Negative Size} & \textbf{Knowledge ID} & \textbf{Total Size} & \textbf{Positive Size} & \textbf{Negative Size} \\ \midrule
\rowcolor[HTML]{EFEFEF} x02030701 & 100 & 25 & 75 & x07020402 & 87 & 29 & 58 \\
x02021101 & 100 & 40 & 60 & x07020502 & 100 & 50 & 50 \\
\rowcolor[HTML]{EFEFEF}x06020104 & 100 & 40 & 60 & x20050401 & 100 & 50 & 50 \\
x02061003 & 100 & 16 & 84 & x09020509 & 100 & 50 & 50 \\
\rowcolor[HTML]{EFEFEF} x48040202 & 100 & 29 & 71 & x07020314 & 100 & 30 & 70 \\
x11041602 & 100 & 24 & 76 & x01010201 & 100 & 50 & 50 \\
\rowcolor[HTML]{EFEFEF} x04030501 & 100 & 48 & 52 & x11040205 & 100 & 26 & 74 \\
x04030601 & 100 & 23 & 77 & x11040203 & 100 & 22 & 78 \\
\rowcolor[HTML]{EFEFEF} x07010103 & 100 & 50 & 50 & x11040202 & 100 & 25 & 75 \\
x06030101 & 100 & 44 & 56 & x02040502 & 100 & 44 & 56 \\
\rowcolor[HTML]{EFEFEF} x57130902 & 100 & 35 & 65 & x47060201 & 100 & 17 & 83 \\
x20041003 & 62 & 50 & 12 & x20070401 & 100 & 47 & 53 \\ \bottomrule
\end{tabular}}
\end{table*}

\subsection{Dataset Overview}
We conduct our experiment with MathKnowCT~\citep{li2024knowledge}, which contains 24 knowledge concepts from math concepts ranging from Grade 1 to Grade 3. The dataset was constructed by finding 100 candidate questions with the highest text embedding similarity to each knowledge concept. For each question, a pedagogical expert annotated whether or not the question fit. The matching to mismatching ratio in the dataset is around 1:4. More details about the knowledge definitions and statistics of the dataset can be found in Tab.~\ref{tab:data_detail} and Tab.~\ref{tab:data_knowledge}. To enhance the performance of LLMs, we randomly sample two examples from each knowledge concept and use them as demonstrations for the few-shot learning implementations.

\subsection{Implement Settings}

To explore the compatibility of our proposed framework, we experiment it with 3 representative LLMs frameworks, including Llama-3~\citep{touvron2023llama}, Mixtral~\citep{jiang2024mixtral}, and GPTs~\citep{brown2020language}. For each frameworks, we choose two sized models, e.g., base and large, to explore the impacts of agent model sizes. In addition, for each LLM framework, we use it for all agents implementations in our framework except for the numerical judger. To ensure the generated code are most reliably executable, we chose OpenAI's GPT-4o (with temperature=0.7) for the numerical judger in all following experiments. For each framework, we experiment with two-sized versions (Base and Large) and the prompt text is adjusted based on the preference of each LLM. We run our experiment with the implementation of huggingface packages\footnote{\url{https://huggingface.co/}} on 8 * Nvidia A100 80G GPUs. The detailed model information are listed in Tab.~\ref{tab:llm_infor}. 

\begin{table}[!btph]
\centering
\caption{LLM implementation with source file links.}
\label{tab:llm_infor}
\resizebox{0.33\textwidth}{!}{
\begin{tabular}{@{}cc@{}}
\toprule
LLM Name & Model ID \\ \midrule
\rowcolor[HTML]{EFEFEF}GPT-Large & gpt-4o-2024-05-13\\
GPT-Base & gpt-3.5-turbo-0125 \\
\rowcolor[HTML]{EFEFEF}Llama3-Large & Llama-3-70B-Instruct \\
Llama3-Base & Llama-3-8B-Instruct \\
\rowcolor[HTML]{EFEFEF}Mixtral-Large & Mixtral-8x7B-Instruct-v0.1 \\
Mixtral-Base & Mistral-7B-Instruct-v0.2 \\
\bottomrule
\end{tabular}}
\end{table}

\noindent Following the prior study~\citep{li2024knowledge}, we evaluate the performance with various metrics including accuracy, precision, recall and F1-score. Specifically, the metrics are calculated with the following formulas: 

\begin{align*}
    &\mathrm{Accuracy = \frac{TP + TN}{TP + FP + TN + FN}} \\
    &\mathrm{Precision = \frac{TP}{TP + FP}},\ \mathrm{Recall = \frac{TP}{TP + FN}} \\
    &\mathrm{F1 = \frac{2 * Precision * Recall}{(Precision + Recall)}}
\end{align*}

\noindent where true positive (TP) samples are the matching knowledge-question pairs successfully discerned, false positive (FP) samples are the unrelated sample pairs misclassified as matching, true negative (TN) are the unrelated pairs correctly filtered, false negative (FN) are the matching pairs dismissed. From an educational perspective, false negatives are often preferable to false positives, as a falsely matched question could disrupt a student’s learning process.

\subsection{Baselines}

\begin{table}[]
\caption{Example knowledge definitions of MathKnowCT}
\label{tab:data_knowledge}
\resizebox{0.49\textwidth}{!}{
\begin{tabular}{p{0.13\textwidth}p{0.41\textwidth}}
\toprule
Knowledge ID & Knowledge Definition \\ \midrule
x01010201 & Learn the definitions of following types of numbers, including integers, odd numbers, even numbers, fractions, decimals, positive numbers, negative numbers, and natural numbers. Common related question types include the following: (1) Select a number of a specified type from a given set of numbers; (2) Determine whether a number is within the defined range; (3) Determine whether a proposition about the classification of numbers is true.  \\ \midrule
x02040502 & Learn the composition of two-digit numbers less than or equal to 100 (how many tens and how many ones). Common related question types include the following: (1) Convert a two-digit number into a combination of tens and ones; (2) Fill in the corresponding two-digit number based on the combination of tens and ones.  \\ \midrule
x02061003 & Learn to use 3 or 4 digits to form a three-digit or four-digit number,   and judge the size relationship between the digits. Related question types are limited to the following: (1) Use 3 digits to form a three-digit number smaller than a certain number. Find the total number of such three-digit numbers, the largest number, and the smallest number. Each digit can only be used once in the combination process. (2) Knowing that the sum of the digits in each digit of a four-digit number is a certain number, find the largest number and the smallest number of this four-digit number. \\ \midrule
x04030501 & Learn to calculate the reciprocal of a number. Common related question types include the following: (1) Calculate the reciprocal of one or more given numbers; (2) Given an equation where the product of a number and a blank is 1, find the value of the number that can be filled in the blank.  \\ \midrule
x48040202 & Learn how to estimate the total purchase price of three items in a shopping scenario. Common related question types include the following: (1)   Given the prices of three items (each item can be a three-digit or two-digit   price), but at least one of the items has a three-digit price, calculate the   approximate total purchase price of the three items; (2) Calculate both the   approximate and exact total purchase price of the three items;  \\ \midrule
x57130902 & Learn to solve feasible combinations by enumeration. Common related question types include the following: (1) Given a numerical value of a total quantity demanded (e.g., total quantity of goods transported, total price),   and the numerical value that each option can provide (e.g., the loading capacity of trucks of different sizes, coins of different denominations),   solve the option combination that just meets the total quantity demanded.   Also note that the numbers in the question stem are all integers, and the numerical value of each option in the combination cannot be wasted (e.g.,   each truck must be fully loaded, and no change is given for the currency). In the problem-solving process, no more than 15 feasible combinations should be   enumerated. \\ \bottomrule
\end{tabular}}
\end{table}

We compare our framework with three representative knowledge tagging frameworks introduced in prior sections, including embedding similarity, pre-trained language fine-tunning and single LLM inference. For each framework, we choose to implement with different high-performance backbone models. Details about each baseline's implementation is shown as follows:  

\begin{itemize}
    \item \textbf{Embedding Similarity:} Two high-performed long text encoding models, sentence-BERT (S-BERT) ~\citep{reimers2019sentence} and text-embedding-3-small\footnote{\url{https://platform.openai.com/docs/guides/embeddings/embedding-models}} are leveraged as the backbone model for the embedding similarity framework. The judgment is determined by the top-$K$ selection on cosine similarity between dense vectors of the encoded knowledge and question text, $x_k$ and $x_q$. 
    
    \item \textbf{PLM Fine-tuning:} Following the prior studies~\citep{huang2023pqsct}, we choose PLMs include BERT~\citep{devlin2018bert}, T5~\citep{raffel2020exploring}, and RoBERTa~\citep{liu2019roberta} as the backbone for our implementation. As the knowledge tagging is formulated as a binary classification task in our paper, we add a binary classification layer to the top of $<$BOS$>$ tokens outputs and fine-tune the parameter of the whole model with the binary entropy loss calculated on the samples in the training set. The learning rate during our fine-tuning process is tuned from 1e-3 to 1e-5.
    
    \item \textbf{Single LLM with 2-shot Inference:} We implement a single LLM with 2-shot inference, following prior work~\citep{li2024knowledge}, which incorporates Chain-of-Thought (COT) instructions into the input prompt. In our implementation, we use three backbone models: Llama3, Mixtral, and GPTs. For simplicity, we employ a random selection strategy for demonstration retrieval.
    
\end{itemize}

\subsection{Result and Discussions}

\begin{table*}[]
\centering
\caption{Comparison between PLM Embedding Similarity, PLM Fine-tune, LLM 2-shot Inference and Multi-Agent LLMs. The best result under the comparable settings is marked with \underline{underline}, and the best result among all settings is marked with \textbf{bold}.}
\label{tab:zero-shot}
\vspace{-3mm}
\resizebox{\textwidth}{!}{
\begin{tabular}{@{}cc|c|cc|ccc|ccc|ccc@{}}
\toprule
\multirow{2}{*}{Metric} & \multirow{2}{*}{\begin{tabular}[c]{@{}c@{}}Model \\ Size\end{tabular}} & \multirow{2}{*}{\begin{tabular}[c]{@{}c@{}}Human\\ Expert\end{tabular}} & \multicolumn{2}{c|}{Embedding Similarity} & \multicolumn{3}{c|}{PLM Fine-tuning} & \multicolumn{3}{c|}{Single LLM} & \multicolumn{3}{c}{Multi-Agent LLMs} \\ \cmidrule(l){4-14} 
 &  &  & GPT-Embed & SBERT & BERT & RoBERTa & T5 & Llama-3 & Mixtral & GPT & Llama-3 & Mixtral & GPT \\ \midrule
\multirow{2}{*}{Accuracy} & Base &  91.75 & 67.43 & \underline{78.90} & 58.45 & 35.51 & \underline{77.18} & 67.08 & \underline{75.42} & 68.85 & 74.91 & \textbf{\underline{79.27}} & 75.40 \\
 & Large & - & - & - & 76.64 & 79.08 & \underline{79.55} & 81.21 & 81.18 & \textbf{\underline{88.38}} & 82.97 & 83.66 & \underline{86.91} \\ \midrule
\multirow{2}{*}{Precision} & Base &  88.86 & 52.68 & \underline{64.66} & 44.03 & 35.51 & \underline{64.70} & 51.84 & \underline{59.74} & 53.31 & 62.01 & \textbf{\underline{68.17}} & 62.28 \\
 & Large & - & - & - & 63.02 & \underline{72.61} & 71.45 & 66.11 & 65.78 & \underline{76.51} & 73.09 & 73.64 & \textbf{\underline{80.47}} \\ \midrule
\multirow{2}{*}{Recall} & Base & 88.16 & 75.27 & \underline{82.39} & 62.77 & \textbf{\underline{100.0}} & 78.63 & 94.62 & 93.15 & \underline{95.30} & 74.57 & \underline{77.42} & 77.02 \\
 & Large & - & - & - & \underline{82.80} & 65.94 & 70.62 & 95.91 & 93.38 & \textbf{\underline{96.77}} & 81.86 & 81.07 & \underline{83.06} \\ \midrule
\multirow{2}{*}{F1-score} & Base &  88.51 & 61.98 & \underline{72.45} & 51.75 & 52.41 & \underline{70.99} & 66.98 & \textbf{\underline{72.80}} & 68.37 & 67.72 & \underline{72.50} & 68.87 \\
 & Large & - & - & - & \underline{71.57} & 69.12 & 71.03 & 78.27 & 77.18 & \textbf{\underline{85.46}} & 77.22 & 77.18 & \underline{81.75} \\ \bottomrule
\end{tabular}}
\end{table*}

Figure~\ref{fig:example} illustrates an example showcasing the outputs of all agents in the system during the inference process for a knowledge-question pair. The performance of the baseline models and our proposed multi-agent system across the entire dataset is presented in Table~\ref{tab:zero-shot}.

\paragraph{RQ1:} 

To address RQ1, we first compare the proposed method with two non-LLM baseline frameworks: embedding similarity and PLM fine-tuning. Our comparison reveals that the base-sized LLMs achieve results comparable to those of the baseline models. As the model size increases, larger LLMs significantly outperform the baselines. When comparing single LLM inference with the multi-agent approach, we find that the introduction of planning and numerical agents leads to substantial improvements in precision. This is because the clear sub-constraints, decomposed from the complex problem definition, reduce false positive errors in predictions. 

However, we also observed a notable decrease in recall with the multi-agent design. This decline can be attributed to errors in the intermediate steps of the multi-step judging process, which increase false negative errors. Apart from that, although RoBERTa-base achieves 100\% recall, its low precision makes it unsuitable for real-world scenarios. Based on these observations, we conclude that the LLM-based multi-agent system is an effective algorithm for the knowledge tagging task.

\paragraph{RQ2:} 

For RQ2, we examine the performance gap between single LLMs and Multi-Agent LLMs across different model sizes. Our analysis reveals that while the multi-agent design significantly improves metrics such as accuracy and precision for base-sized LLMs, the benefits are less pronounced for large-sized LLMs. This suggests that larger LLMs are inherently more capable of handling complex tasks compared to smaller models. However, a closer inspection of precision shows that our proposed multi-agent framework consistently enhances precision, even for large-sized LLMs. Given the educational context, we can still assert that the multi-agent framework adds value to large-sized LLMs in knowledge tagging tasks.

Furthermore, as shown in Tab.~\ref{tab:zero-shot}, the performance gap between base-sized and large-sized LLMs is significantly reduced with the multi-agent approach. Considering the cost-effectiveness of the entire model, we believe that the proposed multi-agent framework has great potential to evolve into a high-performance, cost-efficient solution for knowledge tagging.

%% file: impact.tex
The designed multi-agent LLM knowledge tagging system has been deployed in Squirrel Ai Learning and applied to massive K-12 students from 338 cities of 35 provinces in China, which demonstrated significant impact and proven to generate substantial business value in several areas. This innovative approach ensures that each problem is properly linked to its specific knowledge concept group automatically at scale, which saves tremendous human efforts and a significant economic cost as well. More importantly, the deployment of this multi-agent LLM system has not only realized significant cost savings but also fundamentally enhanced how educational content is created, delivered, and evaluated. In particular, it has extensive impacts on quality control of contents generated by LLMs and directs such  

\subsection{Direct Impact on Cost Savings}
Focusing initially on primary school content, specifically mathematics for grades 1 to 5, this project covers approximately 1,900 knowledge points of the Chinese math program and around 2,100 that of the U.S. math program. This automation greatly reduces the effort of human labeling which has been the primary solution for over a decade. The system was deployed at the starting of summer 2024 and has saved at least \$306,750 labeling cost when 75 unique problems per knowledge point and \$1 per pair of tagging are taken into account.

\subsection{Indirect Impacts on Educational Quality and Efficiency}
The project's indirect impacts significantly enhance the educational experience through smarter content delivery and improved diagnostic tools:

\begin{itemize}
    \item \textbf{Quality Control of Problem Generation:} Lacking sufficient problems has been a pain point for several years and there were no low-cost and time-effective solution until LLMs arise. However, creating problems with LLMs, despite its novelty and efficiency, still relies heavily on experts' efforts to validate them before serving to students. Therefore, the multi-agent knowledge tagging system played a critical role here that it served as a smart judge that tirelessly labeled (problem, knowledge point) pairs, which boosted the efficiency of problem validation by more than 90\%. With the support of LLM problems generation, followed by multi-agent knowledge tagging as a judge, the problems pool has tripled since the deployment of the system.
    \item \textbf{Improved Tag Linking and Recommendation Systems:} Precise links between problems and knowledge points allow for more effective recommendations of content tailored to individual student needs, helping to identify and fill gaps in understanding. With the capacity to generate a larger pool of problems—up to three times more for each knowledge point—the likelihood of students encountering repeated problems has been reduced by 70\%, enhancing learning efficiency and engagement.
    \item \textbf{Enhanced Diagnostic Tools:} The exact mapping of problems to knowledge points also refines the diagnostics of learning errors. When a student makes a mistake, the system can quickly pinpoint the specific knowledge point involved, enabling more accurate and constructive error analysis. This feature has improved the relevance of error reasoning in about 10\% of cases, providing direct, actionable feedback to both students and educators.
\end{itemize}